\documentclass[10pt,twocolumn,letterpaper]{article}

\usepackage{cvpr}
\usepackage{times}
\usepackage{epsfig}
\usepackage{graphicx}
\usepackage{amsmath}
\usepackage{amssymb}

\usepackage{amsmath}
\usepackage{amssymb}
\usepackage{amsthm}
\usepackage{mathtools}
\usepackage{float}
\usepackage[utf8]{inputenc}
\usepackage{float}
\usepackage{subcaption}
\usepackage{textcomp}

\def \cS {\mathcal{S}}
\def \cC {\mathcal{C}}

\def \cD {\mathcal{D}}

\def \cL {\mathcal{L}}
\def \cG {\mathcal{G}}
\def \cT {\mathcal{T}}
\def \RR {\mathbb{R}}

\makeatletter
\newsavebox\myboxA
\newsavebox\myboxB
\newlength\mylenA

\newcommand*\xoverline[2][0.75]{%
    \sbox{\myboxA}{$\m@th#2$}%
    \setbox\myboxB\null
    \ht\myboxB=\ht\myboxA%
    \dp\myboxB=\dp\myboxA%
    \wd\myboxB=#1\wd\myboxA
    \sbox\myboxB{$\m@th\overline{\copy\myboxB}$}
    \setlength\mylenA{\the\wd\myboxA}
    \addtolength\mylenA{-\the\wd\myboxB}%
    \ifdim\wd\myboxB<\wd\myboxA%
       \rlap{\hskip 0.5\mylenA\usebox\myboxB}{\usebox\myboxA}%
    \else
        \hskip -0.5\mylenA\rlap{\usebox\myboxA}{\hskip 0.5\mylenA\usebox\myboxB}%
    \fi}
\makeatother

\usepackage[pagebackref=true,breaklinks=true,letterpaper=true,colorlinks,bookmarks=false]{hyperref}

\cvprfinalcopy 

\def\cvprPaperID{6841} 

\ifcvprfinal\fi
\begin{document}

\title{Adversarial Camouflage: Hiding Physical-World Attacks with Natural Styles}

\author{Ranjie Duan\textsuperscript{1}\ \  
Xingjun Ma\textsuperscript{2}\footnotemark[2] \ \  
Yisen Wang\textsuperscript{3} \ \
James Bailey\textsuperscript{2}\ \
A. K. Qin\textsuperscript{1} \ \ Yun Yang\textsuperscript{1}\\
\textsuperscript{1}Swinburne University of Technology \ \ \textsuperscript{2}The University of Melbourne \ \  \textsuperscript{3}Shanghai Jiao Tong University\\
}

\maketitle
\thispagestyle{empty}

\renewcommand{\thefootnote}{\fnsymbol{footnote}} 
\footnotetext[2]{Correspondence to: Xingjun Ma (xingjun.ma@unimelb.edu.au)} 
\footnotetext[3]{Code is available at https://github.com/RjDuan/AdvCam-Hide-Adv-with-Natural-Styles}
\begin{abstract}
Deep neural networks (DNNs) are known to be vulnerable to adversarial examples. Existing works have mostly focused on either digital adversarial examples created via small and imperceptible perturbations, or physical-world adversarial examples created with large and less realistic distortions that are easily identified by human observers. In this paper, we propose a novel approach, called Adversarial Camouflage (\emph{AdvCam}), to craft and camouflage physical-world adversarial examples into natural styles that appear legitimate to human observers. Specifically, \emph{AdvCam} transfers large adversarial perturbations into customized styles, which are then ``hidden'' on-target object or off-target background. Experimental evaluation shows that, in both digital and physical-world scenarios, adversarial examples crafted by \emph{AdvCam} are well camouflaged and highly stealthy, while remaining effective in fooling state-of-the-art DNN image classifiers. Hence, \emph{AdvCam} is a flexible approach that can help craft stealthy attacks to evaluate the robustness of DNNs. 
\emph{AdvCam} can also be used to protect private information from being detected by deep learning systems.  

\end{abstract}

\section{Introduction}
\begin{figure}[ht]
  \centering
  \includegraphics[width=\linewidth]{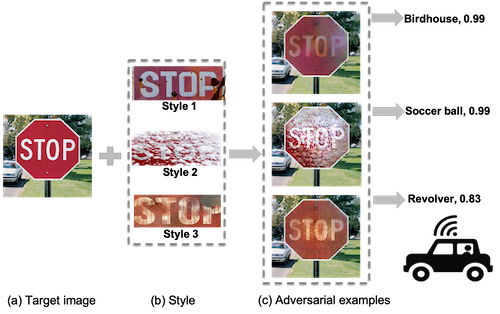}
  \caption{Camouflaged adversarial examples crafted by proposed \emph{AdvCam} attack. Given a target image in (a), an adversary can choose different camouflage styles from (b) to craft adversarial examples in (c) that appear naturally occurring, yet can fool a DNN classifier to make incorrect predictions with high confidence.} 
  \label{fig:stop_sign}
\end{figure}

Deep neural networks (DNNs) are a family of powerful models that have been widely used in various AI systems, and they have achieved a great success across many applications, such as image classification \cite{he2016deep}, 
speech recognition \cite{wang2017residual}, 
natural language processing \cite{zeng2019dirichlet}, 
and autonomous driving \cite{chen2015deepdriving}.
However, DNNs are known to be vulnerable to adversarial examples (or attacks) that are crafted by adding carefully designed perturbations on normal examples \cite{szegedy2013intriguing,goodfellow2014explaining,ma2018characterizing,bai2019hilbert,wang2019convergence,Wang2020Improving}.
This raises serious concerns for security-critical applications \cite{sharif2016accessorize,dong2019efficient,evtimov2017robust,jiang2019black,ma2019understanding}.
For example, as shown in Figure \ref{fig:stop_sign}, the addition of a carefully crafted perturbation that resembles stain, snow and discoloration on the surface of a stop sign. A self-driving car equipped with state-of-the-art classifier detects the modified stop sign as other objects with very high confidence (we will examine how this perturbation was created later).

Adversarial attacks can be applied in two different settings: 1) \textbf{digital setting}, where the attacker can feed input digital images directly into the DNN classifier; and 2) \textbf{physical-world setting}, where the DNN classifier only accepts inputs from a camera and the attacker can only present adversarial images to the camera. There are three properties which may be used to characterize adversarial attack: 1) \textbf{ adversarial strength}, which represents the ability to fool
DNNs; 2) \textbf{adversarial stealthiness}, which is about whether the adversarial perturbations can be detected by human observers; and 3) \textbf{camouflage flexibility}, which is the degree to which the attacker can control the appearance of adversarial image.

Most attacking methods have been developed for the digital setting, such as Projected Gradient Decent (PGD) \cite{madry2017towards}, Carlini and Wagner (CW) attack \cite{carlini2017towards} and adversarial examples crafted using generative adversarial networks (AdvGAN) \cite{xiao2018generating}. For digital attacks, small perturbations are often sufficient. However, physical-world attacks require large or even unrestricted perturbations \cite{kurakin2016adversarial,athalye2017synthesizing}, since small perturbations are too subtle to be captured by cameras in complex physical-world environments. There already exist several attacking methods that go beyond small perturbations, such as adversarial patch (AdvPatch) \cite{brown2017adversarial} and robust physical perturbations ($RP_2$) \cite{evtimov2017robust}. The properties of existing attacks are summarized in Table \ref{tab1}, where $\star \star$ means better than $\star$.
To summarize, stealthiness can be achieved with small perturbations, which are only useful in digital setting.
Also, existing attacks require exact perturbation size to achieve stealthiness, while it is difficult to decide a proper perturbation size for both visual imperceptibility and adversarial strength, especially in the physical setting. Besides, the generation process of current methods is difficult to control, e.g., an attacker cannot decide the appearance of adversarial examples. As such, the camouflage flexibility of these methods is rather limited. A flexible (yet strong and stealthy) camouflage mechanism for large perturbations is still an open problem that needs to be addressed.



\begin{figure}[!t]
  \centering
  \begin{subfigure}[b]{0.24\linewidth}
    \includegraphics[width = \linewidth]{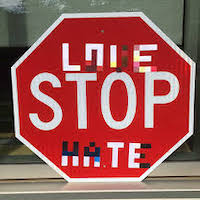}
    \caption{$RP_2$}
  \end{subfigure}
  \begin{subfigure}[b]{0.24\linewidth}
    \includegraphics[width = \linewidth]{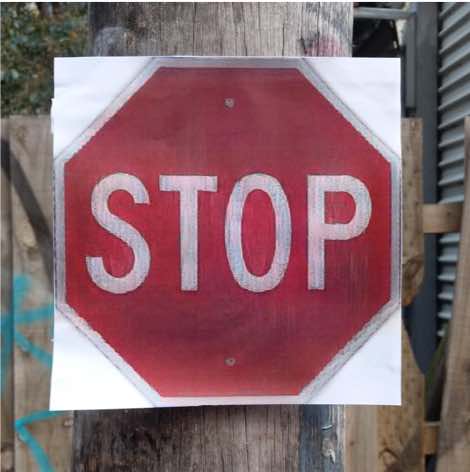}
    \caption{AdvCam}
  \end{subfigure}
  \begin{subfigure}[b]{0.24\linewidth}
    \includegraphics[width = \linewidth]{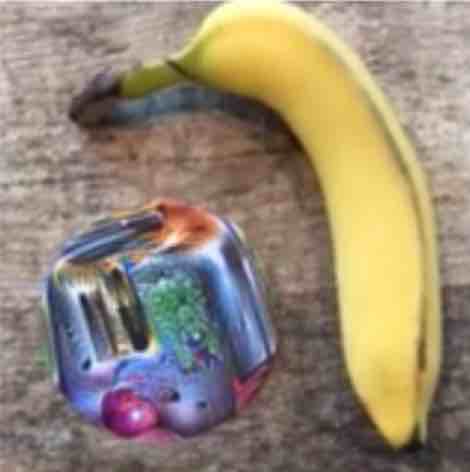}
    \caption{AdvPatch}
  \end{subfigure}
\begin{subfigure}[b]{0.24\linewidth}
    \includegraphics[width = \linewidth]{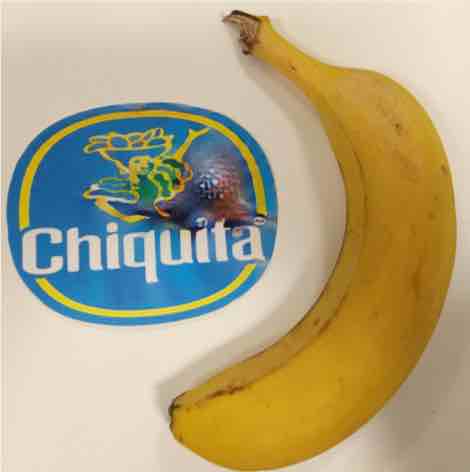}
    \caption{\emph{AdvCam}}
  \end{subfigure}
  \caption{Examples of successful physical-world attacks. \emph{AdvCam} refers to our proposed adversarial camouflage.}
  \label{fig:other_work}
\end{figure}


\vspace{-0.1 in}
\begin{table}[htb]
\centering
\small
\caption{Summary of existing attacks and our \emph{AdvCam}.}
\label{tab1}
\begin{tabular}{lcccc}
\hline
Attack & Digital & Physical & Stealthiness & Flexibility \\ \hline
PGD &  $\surd$ & $\times$ & $\star \star$ & $\star$ \\
AdvPatch & $\times$ & $\surd$ & $\star $ & $\star$ \\
$RP_2$ & $\times$ & $\surd$ & $\star $ & $\star$ \\
\hline
\emph{AdvCam} & $\surd$ & $\surd$ &$\star \star $ & $\star \star $ \\ \hline
\end{tabular}
\end{table}

To address this gap, in this paper, we propose a novel adversarial camouflage approach (\emph{AdvCam}) to craft and camouflage adversarial examples into natural styles using style transfer techniques. 
The style of an image is an abstract concept that generally refers to its visual appearance such as color and texture, in contrast to its structure information \cite{karayev2013recognizing}.
In \emph{AdvCam}, the camouflage style and attack region can be customized by the attacker according to different attacking scenarios. For example, Figure \ref{fig:stop_sign} shows several adversarial traffic signs crafted by our \emph{AdvCam} attack. A quick visual comparison of \emph{AdvCam} to existing physical-world attacks can be found in Figure \ref{fig:other_work}. While all the adversarial examples in Figure \ref{fig:other_work} attack DNNs successfully, we can see that \emph{AdvCam} is able to generate adversaial perturbation on stop sign with natural stains compared to artificial graffiti created by $RP_2$, or a camouflaged product label compared to a patch with obtrusive pattern generated by AdvPatch. Our proposed \emph{AdvCam} is capable of generating highly stealthy adversarial examples, which are robust to various physical-world conditions at the same time.

In summary, \emph{AdvCam} is not a perturbation-restricted attack and therefore is not inherently subjected to the limited amount of perturbation which is typically required in existing perturbation-restricted techniques. We define a flexible mechanism that induces perturbation appearing in natural-looking, which is a totally different paradigm from previous attacks. It is such an intrinsic difference of the working principles that makes \emph{AdvCam} to produce more realistic images than existing methods.

Our key contributions in this paper are:

\begin{itemize}
    \item We propose a flexible adversarial camouflage approach, \emph{AdvCam}, to craft and camouflage adversarial examples.
    
    
    \item \emph{AdvCam} allows the generation of large perturbations, customizable attack regions and camouflage styles. It is very flexible and useful for vulnerability evaluation of DNNs against large perturbations for physical-world attacks.
    
    \item Experiments on both digital and physical-world scenarios show that adversarial examples camouflaged by \emph{AdvCam} are highly stealthy, while remaining effective in fooling state-of-the-art DNN image classifiers.
    
\end{itemize}


\section{Related Work}\label{sec:related_work}

\subsection{Adversarial Attack}
Adversarial attack is to generate adversarial examples by maximizing the classification error of the target model (the model to attack) \cite{szegedy2013intriguing}. There are targeted and untargeted attacks. Targeted attack is to fool the network to misclassify the adversarial example into the class that attacker expects, while untargeted attack is to fool the network to misclassify the adversarial example into any incorrect classes \cite{goodfellow2014explaining}.
Adversarial attacks can be applied either in a digital setting directly on the target model, or in a physical-world setting, where recaptured photos of adversarial examples are fed to the target model \cite{kurakin2016adversarial}.

\subsubsection{Digital attacks}
Adversarial examples can be crafted by one or more steps of perturbation following the direction of adversarial gradients. This includes the classic Fast Gradient Sign Method (FGSM) \cite{goodfellow2014explaining}, the Basic Iterative Method (BIM) \cite{kurakin2016adversarial}, the strongest first-order method Projected Gradient Decent (PGD) \cite{madry2017towards}, and the Skip Gradient Method (SGD) \cite{wu2020skip} for transferable attacks. These attacks can either be targeted or untargeted, and their perturbations are bounded by a small norm-ball $\left\lVert \cdot \right\rVert_p \leq \epsilon$ with $L_2$ and $L_\infty$ being the most commonly used norms. Optimization-based attacks, such as Carlini and Wagner (CW) attack \cite{carlini2017towards} and elastic-net (EAD) attack \cite{chen2018ead}, directly minimize the perturbations as part of the adversarial loss.

There also exist several unrestricted adversarial attacks. These attacks search for modifications on substitutable components (attributes) of an image such as color \cite{hosseini2018semantic}, texture and physical parameters \cite{zeng2019adversarial,liu2018beyond} while preserving critical components of images. However, these attacks either produce large unnatural distortions or rely on training data that has semantic annotations. Moreover, these attacks cannot generate complex adversarial patterns and thus are quite limited for complicated real scenarios. Adversarial examples can also be generated by generative adversarial networks (GANs) \cite{xiao2018generating,song2018constructing}. 
However, it is difficult to craft targeted attacks for a given test image with GANs, since the generation process is hard to control.


Most existing attacks achieve stealthiness by either crafting small perturbations or modifying semantic attributes of the target image. However, a flexible camouflage mechanism that can effectively hide adversarial examples with natural styles is still missing from the literature. In this paper, we address this gap by proposing one such approach.

\subsubsection{Physical-world attacks}
A study has shown that, by printing and recapturing using a cell-phone camera, digital adversarial examples can still be effective \cite{kurakin2016adversarial}. However, follow-up works have found that such attacks are not easy to realize under physical-world conditions, due to viewpoint shifts, camera noise, and other natural transformations \cite{athalye2017synthesizing}. Thus, strong physical-world attacks require large perturbations, and specific adaptations over a distribution of transformations including lighting, rotation, perspective projection etc. The AdvPatch attack allows large perturbations and is immune to scaling or rotation, thus being directly applicable as a physical-world attack \cite{brown2017adversarial}. Adversarial stickers and graffiti have also been used to attack such as traffic sign classifiers and ImageNet classifiers in physical-world scenarios \cite{evtimov2017robust}. Other physical-world attacks include adversarial eye-glass frames \cite{sharif2016accessorize}, vehicles \cite{zhang2018camou}, or t-shirts \cite{xu2019adversarial} that can fool face recognition systems or object detectors. All these physical-world attacks generate large perturbations to increase adversarial strength, which inevitably results in large and unrealistic distortions. This greatly reduces their stealthiness, as shown in Figure \ref{fig:other_work}.

\subsection{Neural Style Transfer}
Neural style transfer evolves from the problem of texture transfer, for which the goal is to transfer the texture of a source image to a target image while preserving the structural information of the target image. Traditional texture transfer methods mainly focus on non-parametric algorithms that resample pixels from the source texture \cite{efros2001image}. However, these methods suffer from the fundamental limitation of pixel replacement, i.e., they cannot handle complex styles. Neural style transfer demonstrates remarkable results for image stylization \cite{gatys2016image}. In it, the content and style information of an image can be separated from its feature representations learned by a convolutional neural network (CNN). Then, the style information of the style image can be recombined into the target image to achieve style transfer. This technique has attracted several follow-up works for different aspects of improvement \cite{champandard2016semantic,luan2017deep}. In this paper, we will exploit these techniques for the camouflage of adversarial examples.



\begin{figure*}[!htb]
  \centering
  \includegraphics[width=0.9\linewidth]{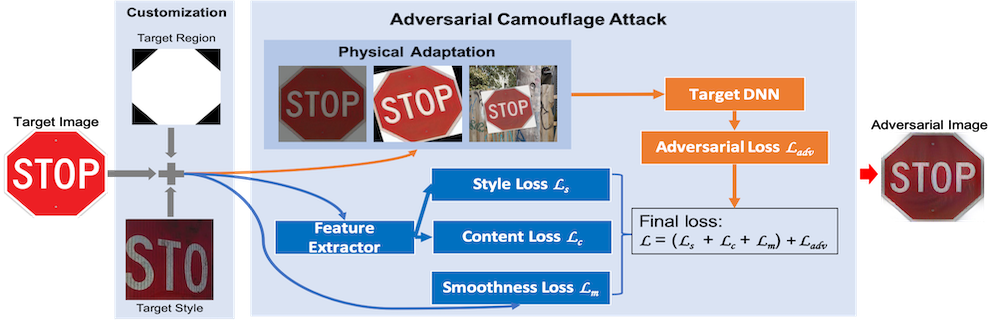}
  \caption{Overview of the proposed approach.}
  \label{fig:framework}
\end{figure*}
\section{Our Adversarial Camouflage Approach}\label{sec:proposed}
In this section, we first give an overview of the adversarial attack problem and our proposed camouflage approach. We then introduce the loss functions used by our proposed approach, and adaptations for physical world conditions.

\subsection{Overview}
Given a test image $x \in \RR^m$ with class label $y$, a DNN classifier $F: \RR^m \to \{1, \cdots, k\}$ mapping image pixels to a discrete
label set, and a target class $y_{adv} \neq y$, adversarial attack is to find an adversarial example $x'$ for target image $x$ by solving the following optimization problem:
\begin{equation}\label{eq:1}
\begin{aligned}
\textrm{minimize} \quad &\cD(x, x') +\lambda \cdot \cL_{adv}(x')\\
\textrm{such that} \quad & x' \in[0, 255]^m,
\end{aligned}
\end{equation}
where $\cD(x, x')$ is a distance metric that defines the stealthiness property of the adversarial example, $\cL_{adv}$ is the adversarial loss, $[0,255]$ indicates the valid pixel values, $\lambda$ is a parameter that adjusts the adversarial strength. Note that there is a trade-off between stealthiness and adversarial strength. In the whole experiments, we fix all other parameters as constants \cite{luan2017deep}, and only adjust adversarial strength parameter $\lambda$.

Our goal is to develop a mechanism that crafts and camouflages adversarial examples with large perturbations into customized styles, and either attack area or style can be defined by attacker flexibly. We use style transfer techniques to achieve the goal of camouflage and adversarial attack techniques to achieve adversarial strength.
The final loss is a combination of an adversarial loss $\cL_{adv}$ for adversarial strength, a style loss $\cL_{s}$ for style generation, a content loss $\cL_c$ to preserve the content of the source image and a smoothness loss $\cL_m$ to generate locally smooth regions. We denote this final loss as the \emph{adversarial camouflage loss}:
\begin{equation}\label{eq:total}
    \cL = (\cL_s + \cL_c + \cL_m) + \lambda \cdot \cL_{adv},
\end{equation}
where the three loss functions in brackets together serve the purpose of camouflage. The overview of our approach is illustrated in Figure \ref{fig:framework}. An attacker defines target image, target attack region and expected target style. Our proposed \emph{AdvCam} then generates adversarial perturbation with expected style on the expected area as shown on the right of Figure \ref{fig:framework}. To make adverarial example robust to various environment conditions, including lighting, rotation, etc., we add an extra physical adaptation training for generated $x'$ in each step.



\subsection{Adversarial Camouflage Loss}

\subsubsection{Style loss}\label{sec:region_attack}
For traditional attacks, the stealthiness metric is defined by $\cD(x, x')=\left\| x -x' \right\|_{p}$, where $\left\| \cdot \right\|_{p}$ is the $L_p$ norm and $L_2$ and $L_\infty$ are typically used. This is to constrain the perturbations to be ``small''. For our proposed camouflage, the stealthiness is defined by a style metric between adversarial example $x'$ and a style reference image $x^s$. The style distance between two images can be defined by their differences in style representations:
\begin{equation}\label{eq:2}
\cD_s = \sum_{l \in \cS_l }\left\|\cG(\widetilde{F}_l(x^s))-\cG(\widetilde{F}_l(x'))\right\|_2^2,
\end{equation}
where $\widetilde{F}$ is a feature extractor (such as a public DNN model) that can be different from the target model, and $\cG$ is the Gram matrix of deep features extracted at a set of style layers of $\widetilde{F}$  \cite{gatys2016image}. As different styles can be learned at different layers, we use all convolutional layers of the network as the style layers. To generate stylized perturbation in expected area of the target image, we denote the masks that define the attack and non-attack regions by $M$ and $\xoverline{M}$ respectively. After each step in generation process, we mask the adversarial image $x'$ by $M$ to make only attack region modifiable with non-attack area unchanged ($x$ masked by $\xoverline{M}$).

\subsubsection{Content loss}
The above style loss can craft an adversarial image in the reference style, however, the content of the adversarial image may appear very different to that of the original image.
The content of the original image can be preserved by a content preservation loss:
\begin{equation}\label{eq:4}
\cL_c =\sum_{l \in \cC_l} \left\|\widetilde{F}_{l}(x)-\widetilde{F}_{l}(x')\right\|_2^2,
\end{equation}
where $\cC_l$ represents the set of content layers used for extracting content representations. This is to ensure that the adversarial image has very similar content to the original image in the deep representation space. We use deeper layers of the feature extractor network as the content layers.

Note that the content loss is optional when the attack only occurs in a small region that does not contain any particular content. However, if the attack region contains semantic content, the content loss can help reduce the semantic difference between the adversarial image and its original version.

\subsubsection{Smoothness loss}
The smoothness of the adversarial image can be improved by reducing the variations between adjacent pixels.
For an adversarial image $x'$, the smoothness loss is defined as:
\begin{equation}\label{6}
\mathcal{L}_{m} = \sum((x'_{i, j}-x_{i+1, j})^{2}+(x'_{i, j}-x_{i, j+1})^{2})^{\frac{1}{2}},
\end{equation}
where $x'_{i,j}$ is the pixel at coordinate $(i,j)$ of image $x'$. Intuitively, this will encourage the image to have low-variance (e.g. smooth) local patches. We note that the smoothness loss is limited in improving stealthiness if the surface of both the target image and the style image are already smooth. But we still recommend adding it in physical setting, as Sharif et al. pointed out \cite{sharif2016accessorize}, the smoothness term is useful for improving the robustness of adversarial examples in physical environment.


\subsubsection{Adversarial loss}
For the adversarial loss $\cL_{adv}$, we use the following cross-entropy loss:
\begin{equation}\label{eq:3}
  \begin{cases}
    \log(p_{y}(x')), &\text{for untargeted attack},\\
    -\log(p_{y_{adv}}(x')) + \log(p_{y}(x')), &\text{for targeted attack}\\
    \end{cases}
\end{equation}
where $p_{y_{adv}}()$ is the probability output (softmax on logits) of the target model $F$ with respect to class $y_{adv}$. 
We note that the proposed camouflage attack is not restricted to the particular form of the adversarial loss, and can be used in combination with existing attacking methods.

\subsection{Adaptation for Physical-world Conditions}
To make adversaries generated by \emph{AdvCam} physically realizable, we model physical conditions in the process of generating camouflaged adversaries.
As the physical-world environment often involves condition fluctuations such as viewpoint shifts, camera noise, and other natural transformations \cite{athalye2017synthesizing}, we use a series of adaptations to accommodate such varying conditions.
In particular, we adapt a technique  similar to  Expectation Over Transformation (EOT) \cite{athalye2017synthesizing} but without expectation. In Xie's work \cite{xie2019improving}, they also adapted EOT to improve the transferability of adversarial examples. However, we aim to improve the adaptation of adversarial examples under various physical conditions. Thus we consider the transformations for simulation of physical-world condition fluctuations, including rotation, scale resize, color shift (to simulate lightening change) and random background. 
\begin{equation}
\min_{x'} \big((\cL_s + \cL_c + \cL_m) + \max_{T \in \cT} \lambda \cdot \cL_{adv}(o + T(x')\big),
\end{equation}
where $o$ represents a random background image that we sample in the physical world, and $T$ represents a random transformation of rotation, resize and color shift.

In principle, ``vision" is the main sense of perception for both human observers and DNN models. By adjusting the style reference image $x^s$ according to the original image $x$ and the background image $o$, the proposed camouflage attack can craft highly stealthy adversarial examples that can deceive both human observers and DNN models. 

\section{Experimental Evaluation}\label{sec:experiments}
\begin{figure*}
  \centering
    \begin{subfigure}[b]{0.10\linewidth}
    \includegraphics[width = \linewidth]{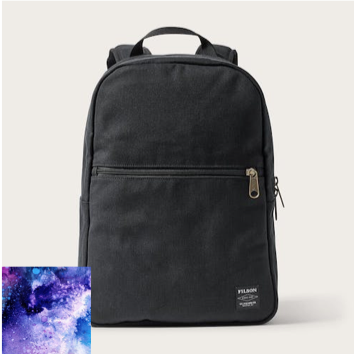}
  \end{subfigure}
    \begin{subfigure}[b]{0.20\linewidth}
    \includegraphics[width = \linewidth]{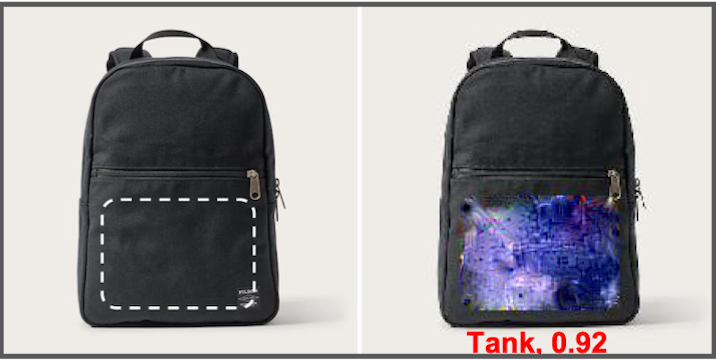}
  \end{subfigure}
    \begin{subfigure}[b]{0.20\linewidth}
    \includegraphics[width = \linewidth]{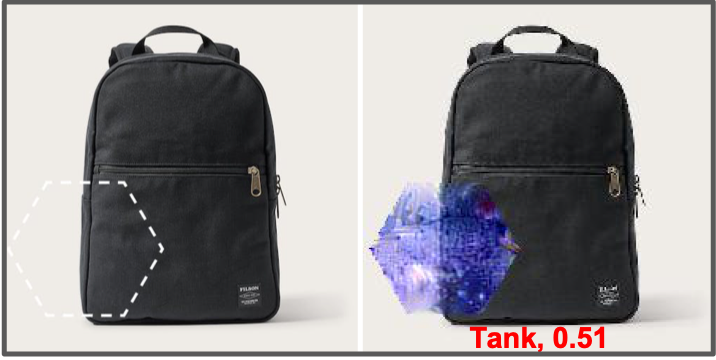}
  \end{subfigure}
    \begin{subfigure}[b]{0.20\linewidth}
    \includegraphics[width = \linewidth]{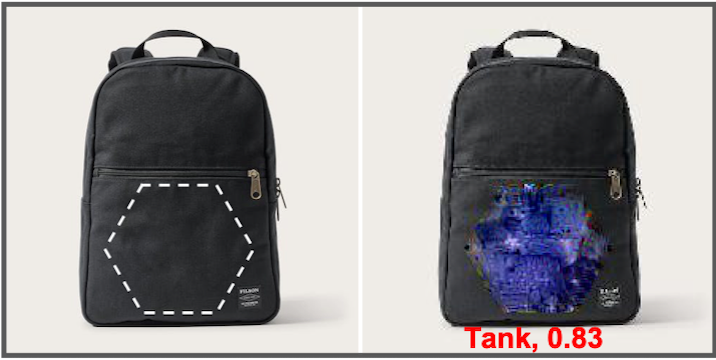}
  \end{subfigure}
    \begin{subfigure}[b]{0.20\linewidth}
    \includegraphics[width = \linewidth]{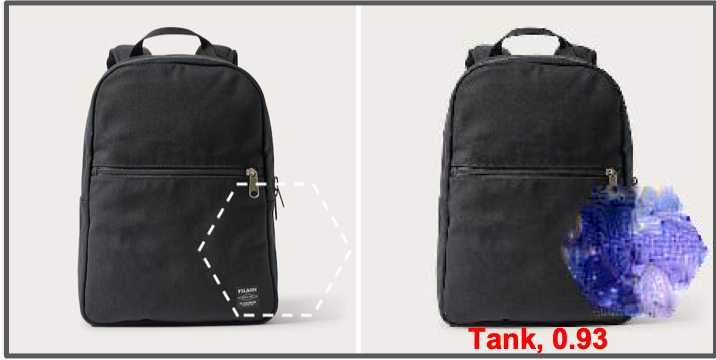}
  \end{subfigure}
    \begin{subfigure}[b]{0.10\linewidth}
    \includegraphics[width = \linewidth]{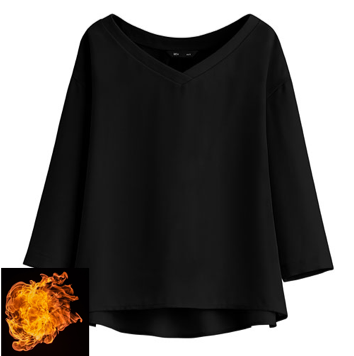}
    \caption{}
  \end{subfigure}
    \begin{subfigure}[b]{0.20\linewidth}
    \includegraphics[width = \linewidth]{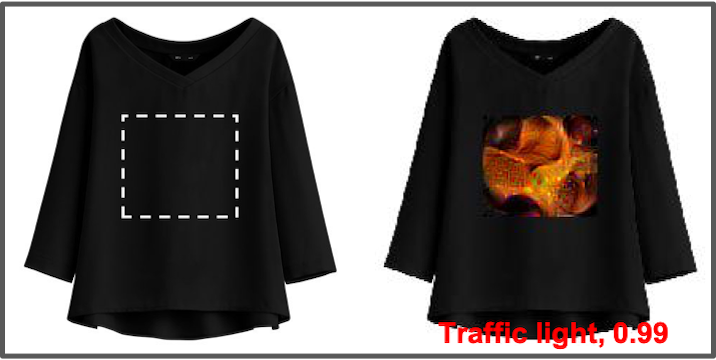}
    \caption{}
  \end{subfigure}
    \begin{subfigure}[b]{0.20\linewidth}
    \includegraphics[width = \linewidth]{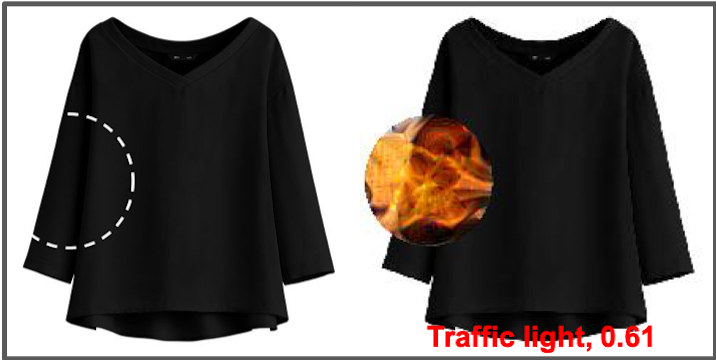}
    \caption{}
    \end{subfigure}
    \begin{subfigure}[b]{0.20\linewidth}
    \includegraphics[width = \linewidth]{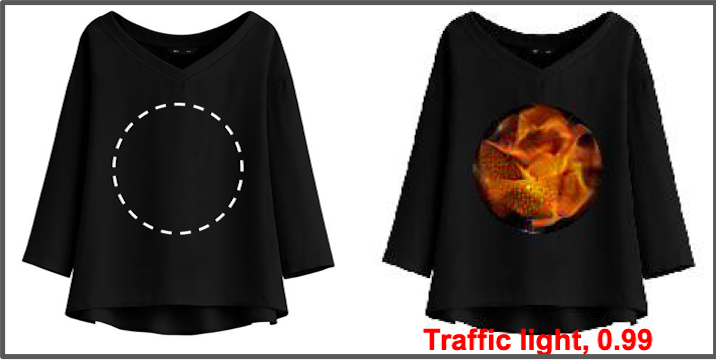}
    \caption{}
    \end{subfigure}
    \begin{subfigure}[b]{0.199\linewidth}
    \includegraphics[width = \linewidth]{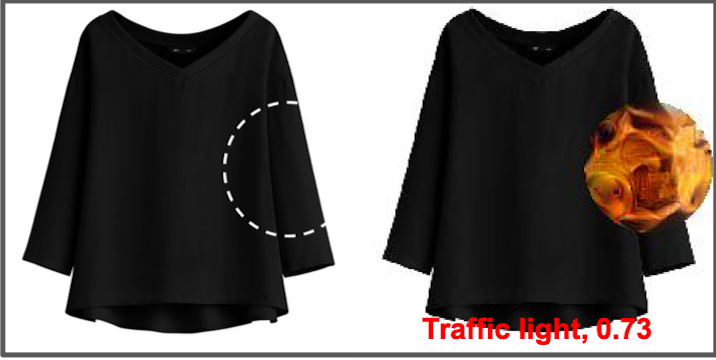}
    \caption{}
  \end{subfigure}
    \caption{Ablation of region shape and size, on two targeted attacks: ``backpack'' $\to$ ``tank'' (top row) and ``poncho'' $\to$ ``traffic light''. (a): original clean image with intended style (bottom left corner). (b) - (e): left: selected attack region, right: crafted camouflage adversarial example with top-1 prediction (``predicted class, confidence'') given at the bottom of the image.}
    \label{fig:various_scales}
\end{figure*}

In this section, we first outline the experimental setup. Then we analyze our \emph{AdvCam} attack via an ablation study. Afterwords, we evaluate camouflage performance of AdvCam by human perception study for digital attacks, and also present several adversarial examples of high stealthiness crafted by \emph{AdvCam}. We further perform physical-world attacks at last.

\subsection{Experimental Setup}
\subsubsection{Baseline attacks} we compare our \emph{AdvCam} attack with two existing representative methods: PGD \cite{madry2017towards} and adversarial patch (AdvPatch) \cite{brown2017adversarial}. PGD represents the digital attacks, which is the strongest first-order attack. And AdvPatch represents the unrestricted attacks that can directly applied to the physical-world setting. Some other physical-world attacks such as $RP_2$ require case-by-case manual design, thus are limited for mass production. 
We compare the methods in terms of attack success rate and visual effect. For \emph{AdvCam} attack, we use the same layers of the network as used in \cite{luan2017deep} to extract style and content (where necessary) representations.

\subsubsection{Threat model}
We test both targeted and untargeted attacks in both digital and physical-world settings. The threat model adopts a gray-box setting: the source and target networks are both VGG-19 networks but were trained separately on ImageNet. For a physical-world test, we first use a Google Pixel2 smartphone to take a photo of the target object, then craft an adversarial image in the digital setting attacking the source network, after which we print out the adversarial image to replace or place it on the target object, then we re-photo the object using the same smartphone from various viewing angles and distances. The physical-word attack success rate is measured by the prediction accuracy of the target network on the re-photoed adversarial images.

\subsection{Ablation Study}
Here, we conduct a series of experiments on ImageNet to analyze the following aspects of our \emph{AdvCam} attack: 1) shape and location of camouflage region, 2) camouflage losses (e.g. style loss, content loss and smoothness loss), and 3) adversarial strength parameter $\lambda$ and region size.

\subsubsection{Camouflage region: shape and location}
Here, we show how the selection of camouflage region in terms of shape and location impacts the adversarial strength of the crafted adversarial example. Given a selected attacking region with shape and size, we increase strength parameter $\lambda$ from 1000 to 10000 with interval 1000 until the attack succeeds. The range is selected according to extensive experiments, [1000, 10000] is an effective range to find adversary with both high adversarial strength and stealthiness. Figure \ref{fig:various_scales} shows camouflaged adversarial examples crafted with different regions. We find that either the camouflage region is at the center or away from the center of the target object, attacks can succeed with high confidence. We will show in the last part of this subsection that attacks can be camouflaged into an area that is even off the target object, secretly hidden in the background.
\begin{figure}
\centering
\begin{subfigure}[b]{0.22\linewidth}
    \includegraphics[width = \linewidth]{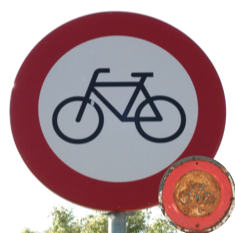}
  \end{subfigure}
\begin{subfigure}[b]{0.22\linewidth}
    \includegraphics[width = \linewidth]{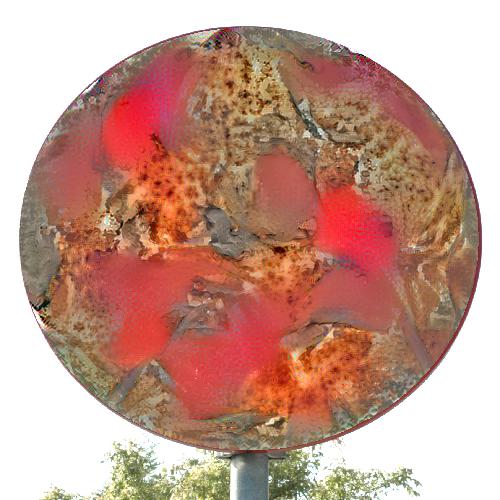}
  \end{subfigure}
\begin{subfigure}[b]{0.22\linewidth}
    \includegraphics[width = \linewidth]{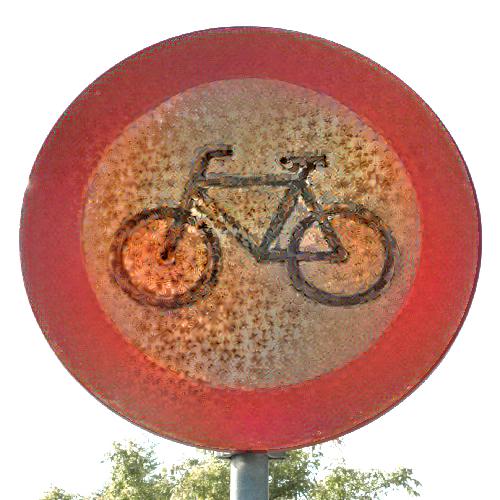}
  \end{subfigure}
\begin{subfigure}[b]{0.22\linewidth}
    \includegraphics[width = \linewidth]{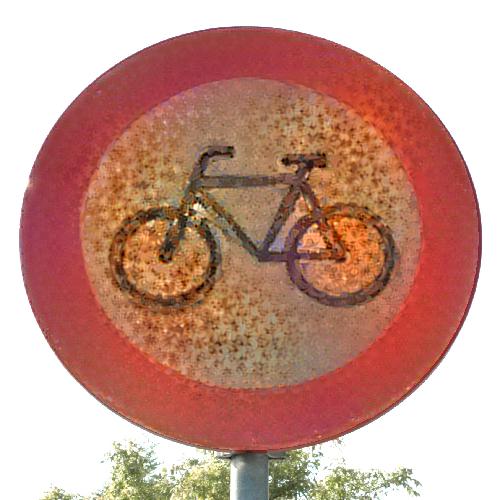}
  \end{subfigure}
\begin{subfigure}[b]{0.22\linewidth}
    \includegraphics[width = \linewidth]{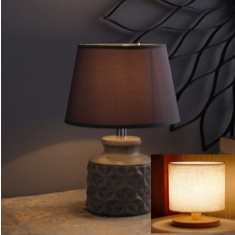}
    \caption{Original}
  \end{subfigure}
\begin{subfigure}[b]{0.22\linewidth}
    \includegraphics[width = \linewidth]{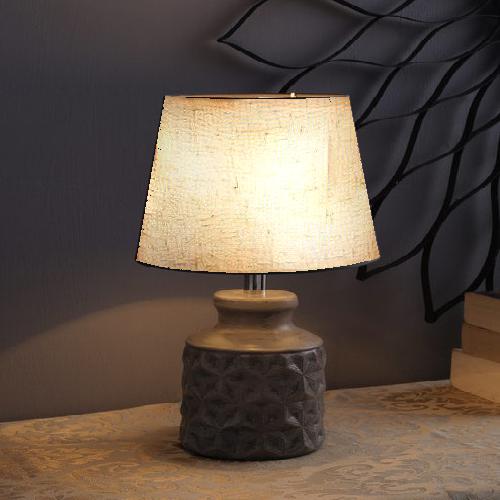}
     \caption{\small{$\cL_s$}}
  \end{subfigure}
\begin{subfigure}[b]{0.22\linewidth}
    \includegraphics[width = \linewidth]{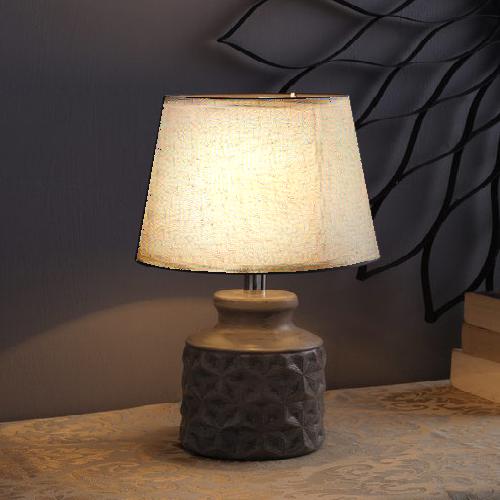}
     \caption{\small{$\cL_s$+$\cL_c$}}
  \end{subfigure}
    \begin{subfigure}[b]{0.22\linewidth}
    \includegraphics[width = \linewidth]{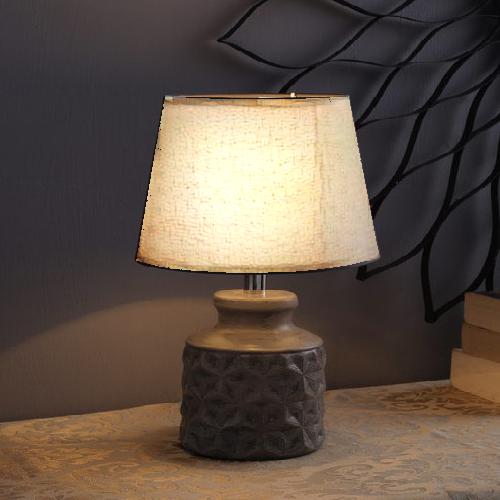}
     \caption{All}
  \end{subfigure}
\caption{Ablation of the 3 camouflage losses: 
(a): original images with intended camouflage style at the bottom right corner; (b) - (d): camouflaged adversarial examples using different loss functions.}
\label{fig:adaptive_constraints}
\end{figure}

\subsubsection{Camouflage losses ($\cL_s$, $\cL_c$, $\cL_m$)}
Figure \ref{fig:adaptive_constraints} illustrates three groups of adversarial examples camouflaged with or without the two optional enhancements (content preservation $\cL_c$ and smoothness enhancement $\cL_m$). When incorporating one enhancement, its loss function is directly added to the final object by following Eq. \eqref{eq:total} ($\lambda$ in $\cL_{adv}$ was set to 2000). As can be observed, the content preservation can help preserve the original content as in the ``traffic sign'' example (third column), while smoothness enhancement can help produce a smooth object surface. These enhancements are optional because they improve the visual appearance only slightly for some adversarial examples, for example, content preservation in the ``table lamp''example (third row) or smoothness enhancement in all examples. 
\begin{figure}[t]
  \centering
  \begin{subfigure}[b]{\linewidth}
    \includegraphics[width = \linewidth]{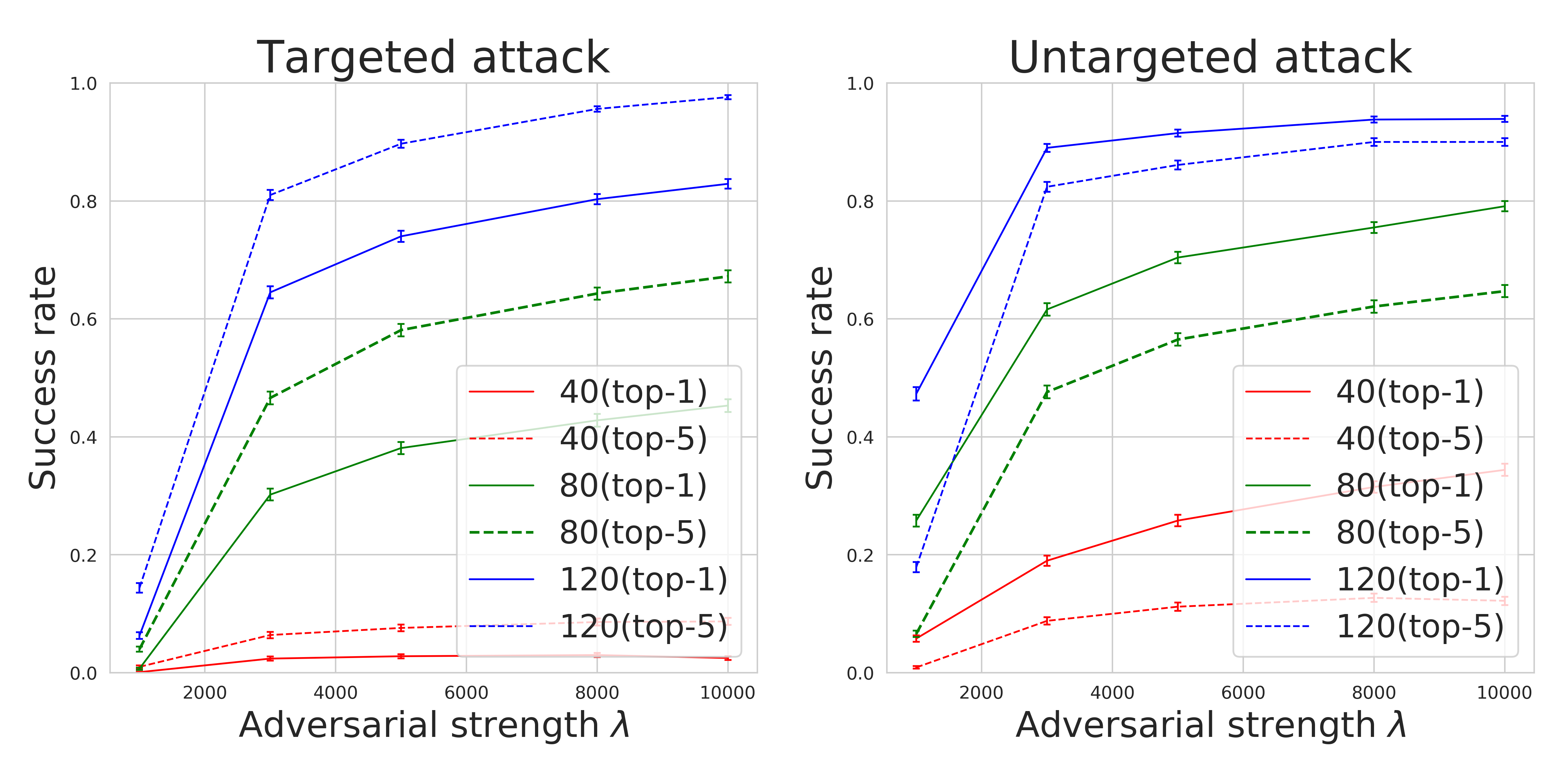}
  \end{subfigure}
    \caption{Ablation of adversarial strength $\lambda$ and region size: success rate of untargeted (left) and targeted attack (right).}
    \label{fig:target_untarget_result}
\end{figure}

\subsubsection{Adversarial strength ($\lambda$) and region size}
We craft both targeted and untargeted camouflage attacks on randomly selected 2000 ImageNet test images from 50 categories with varying $\lambda \in [1000, 10000]$. For targeted attack, the target class is randomly selected and is different to the true class. To also test the influence of $\lambda$ under different attack regions, here we also vary the size of the region from 40*40 to 120*120.
Figure \ref{fig:target_untarget_result} illustrates the top-1/5 success rates, which are measured by whether the target class is in the top-1/5 classes. As shown in the figure, when the region is fixed, larger adversarial strength $\lambda$ can increase success rate by up to 20\%; and when $\lambda$ is fixed, larger attack region can improve success rate by up to 40\%. Compared to targeted attacks, untargeted attacks are much easier to succeed. Between top-1 and top-5 success rates, top-1 targeted attacks are more difficult to achieve (solid lines are lower than dashed lines). The standard errors with respect to different adversarial strengths and region sizes are between 0.07\% and 1.13\%. Note that in these experiments, the camouflage styles and locations of attack region are selected randomly, which may decrease the camouflage effect and success rate. However, this does not affect the conclusion that larger $\lambda$ and region size can help craft stronger attacks.

\subsection{Digital Attacks}\label{sec:comparison}
\subsubsection{Attack setting}
\begin{figure}[h]
  \centering
  \begin{subfigure}[b]{0.24\linewidth}
    \includegraphics[width = \linewidth]{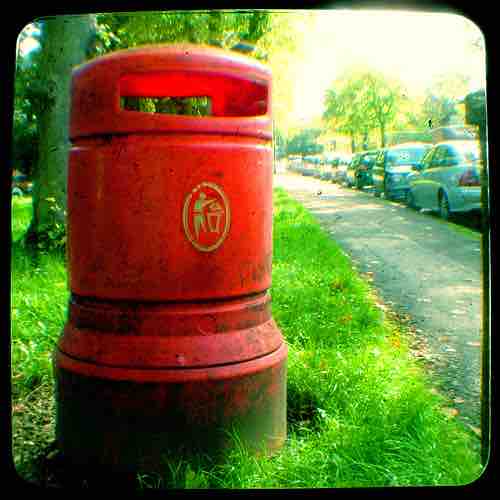}
  \end{subfigure}
  \begin{subfigure}[b]{0.24\linewidth}
    \includegraphics[width = \linewidth]{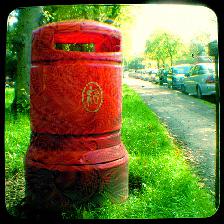}
  \end{subfigure}
    \begin{subfigure}[b]{0.24\linewidth}
    \includegraphics[width = \linewidth]{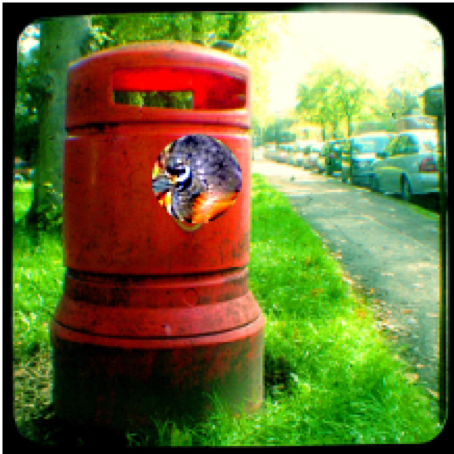}
  \end{subfigure}
  \begin{subfigure}[b]{0.24\linewidth}
    \includegraphics[width = \linewidth]{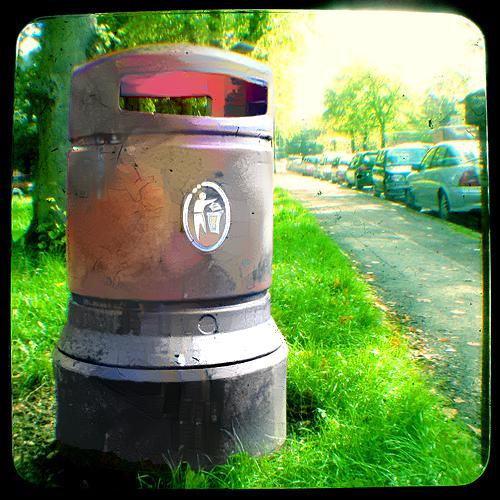}
  \end{subfigure}
 \begin{subfigure}[b]{0.24\linewidth}
    \includegraphics[width = \linewidth]{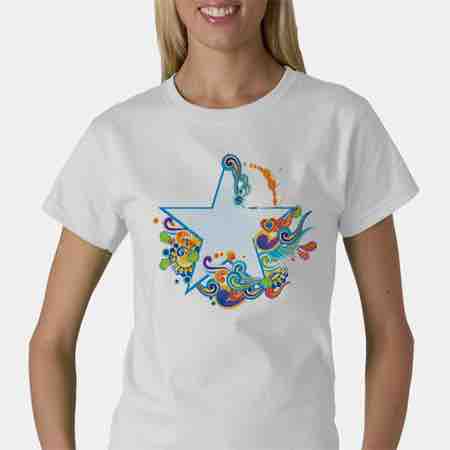}
     \caption{Original}
  \end{subfigure}
  \begin{subfigure}[b]{0.24\linewidth}
    \includegraphics[width = \linewidth]{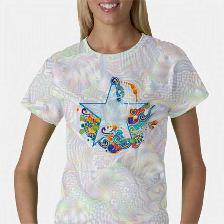}
    \caption{PGD-16}
  \end{subfigure}
    \begin{subfigure}[b]{0.24\linewidth}
    \includegraphics[width = \linewidth]{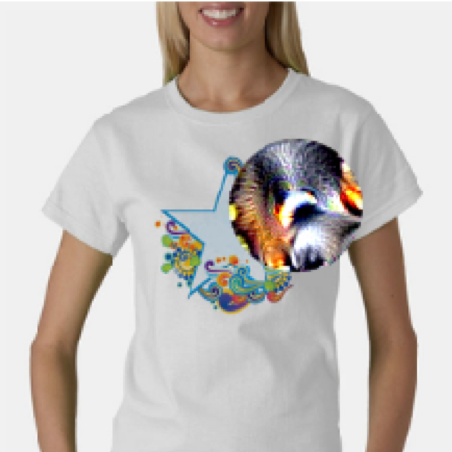}
      \caption{AdvPatch}
  \end{subfigure}
   \begin{subfigure}[b]{0.24\linewidth}
    \includegraphics[width = \linewidth]{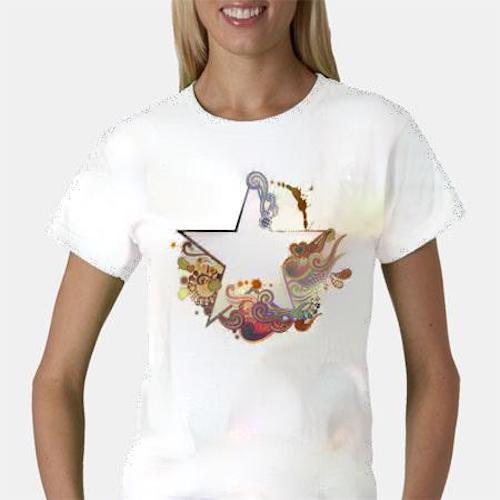} 
    \caption{\emph{AdvCam}}
  \end{subfigure}
  \caption{The original and adversarial images crafted by PGD-16, AdvPatch and our \emph{AdvCam} attack.}
  \label{fig:compare_method_imgs}
\end{figure}

We randomly select 150 clean images from 5 categories of ImageNet ILSVRC2012 test set. We then apply the three methods (PGD, AdvPatch and our \emph{AdvCam}) to craft a targeted adversarial example for each clean image. The pairs of selected source and target classes are: ``ashcan'' $\to$ ``robin'', ``backpack'' $\to$ ``water jug'', ``cannon'' $\to$ ``folklift'', ``jersey'' $\to$ ``horizontal bar'', ``mailbox'' $\to$ ``entertainment center''. For PGD and \emph{AdvCam}, we attack the main object region obtained via manual selection, while for AdvPatch, we further select a circular attack area inside of the object region.
For PGD, we use maximum perturbation $\epsilon=16/255$ (denoted as PGD-16). For \emph{AdvCam}, we randomly select a second image from the same category as the style image, and gradually increase $\lambda$ from 1000 to 10000 until an adversarial example is found. 
For fair comparison, we filter out the failed adversarial examples.
Finally, we collect 132, 101, 122 adversarial examples for PGD, AdvPatch and \emph{AdvCam} respectively. Figure \ref{fig:compare_method_imgs} shows a few crafted adversarial examples by the three methods that we used to perform human perception study.

\subsubsection{Human perception study results}
\begin{figure}[h]
  \centering
  \includegraphics[width=0.9\linewidth]{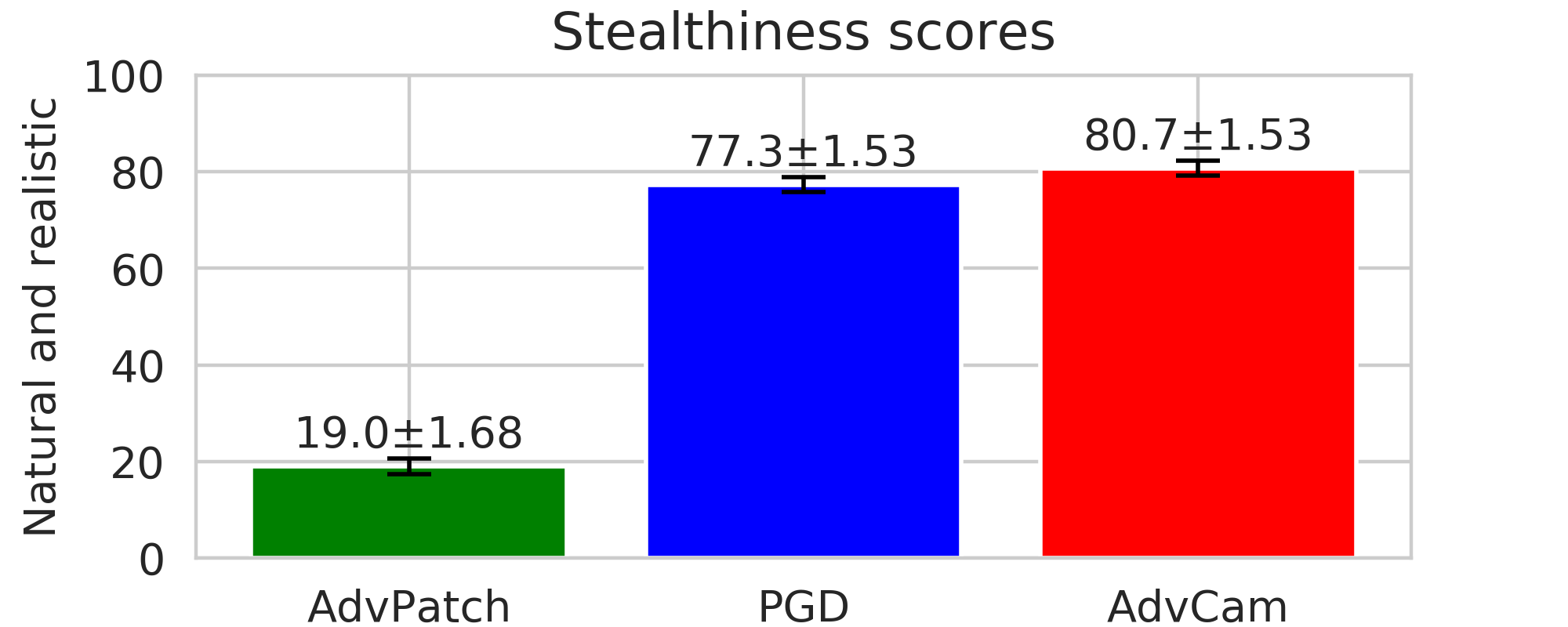}
  \caption{Stealthiness of AdvPatch, PGD-16 and $\emph{AdvCam}$.} 
  \label{fig:stealthiness_scores}
\end{figure}

We set up a human perception study on Amazon Mechanical Turk (AMT) to ask human evaluators to choose whether a shown image is ``natural and realistic'' or ``not natural or realistic''. To simulate adversarial examples in a real world scenario, we present users with three types of adversarial images in random order and individually rather than in pairs. Finally, we collected 1953 selections from 130 participants. AdvPatch was chosen as ``natural and realistic'' $19.0 \pm 1.68\%$ of the time, PGD was chosen $77.3 \pm 1.53\%$ of the time, while our \emph{AdvCam} was chosen $80.7 \pm 1.53\%$ of the time. We summarize these statistics as stealthiness scores for the three methods and show in Figure \ref{fig:stealthiness_scores}. This confirms that our proposed \emph{AdvCam} attack is capable of crafting adversarial examples that are as stealthy as the small perturbation PGD-16 method, although the perturbations of \emph{AdvCam} attacks are unrestricted in size.

\begin{figure*}[t]
  \centering
    \begin{subfigure}[b]{0.16\linewidth}
    \includegraphics[width = \linewidth]{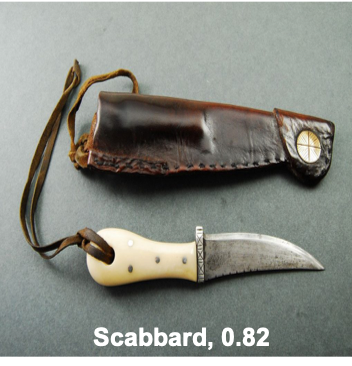}
  \end{subfigure}
    \begin{subfigure}[b]{0.16\linewidth}
    \includegraphics[width = \linewidth]{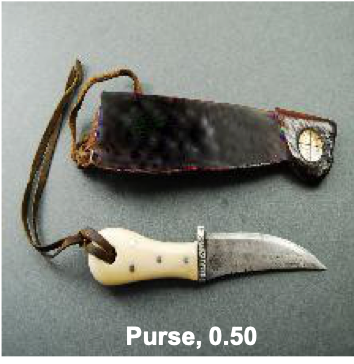}
  \end{subfigure}
    \begin{subfigure}[b]{0.16\linewidth}
    \includegraphics[width = \linewidth]{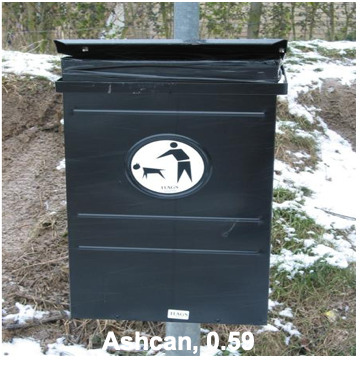}
  \end{subfigure}
    \begin{subfigure}[b]{0.16\linewidth}
    \includegraphics[width = \linewidth]{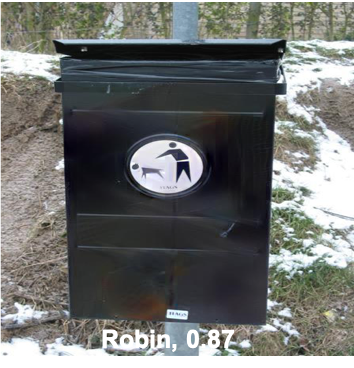}
  \end{subfigure}
    \begin{subfigure}[b]{0.16\linewidth}
    \includegraphics[width = \linewidth]{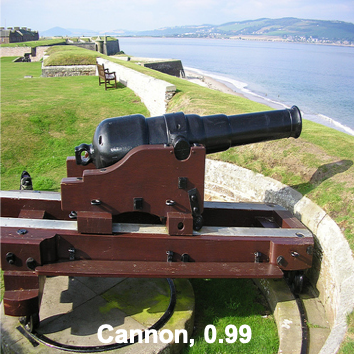}
  \end{subfigure}
    \begin{subfigure}[b]{0.16\linewidth}
    \includegraphics[width = \linewidth]{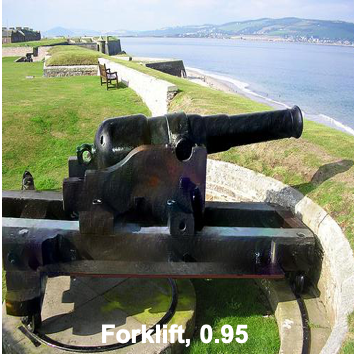}
  \end{subfigure}

    \begin{subfigure}[b]{0.16\linewidth}
    \includegraphics[width = \linewidth]{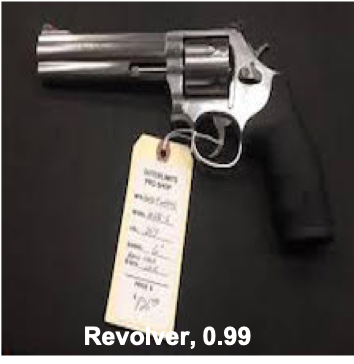}
  \end{subfigure}
    \begin{subfigure}[b]{0.16\linewidth}
    \includegraphics[width = \linewidth]{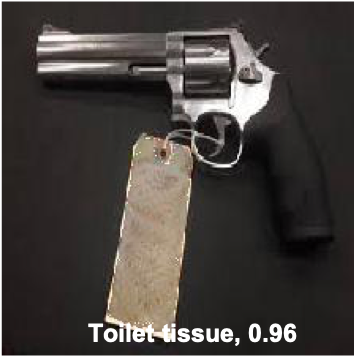}
  \end{subfigure}
    \begin{subfigure}[b]{0.16\linewidth}
    \includegraphics[width = \linewidth]{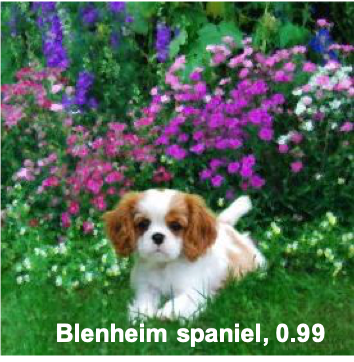}
  \end{subfigure}
    \begin{subfigure}[b]{0.16\linewidth}
    \includegraphics[width = \linewidth]{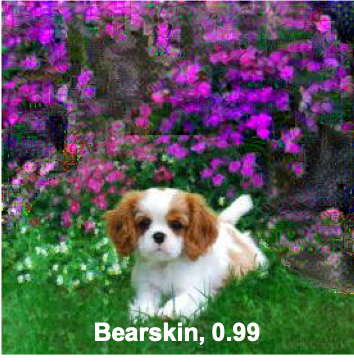}
  \end{subfigure}
    \begin{subfigure}[b]{0.16\linewidth}
    \includegraphics[width = \linewidth]{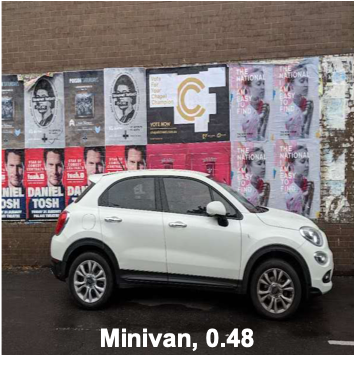}
  \end{subfigure}
    \begin{subfigure}[b]{0.16\linewidth}
    \includegraphics[width = \linewidth]{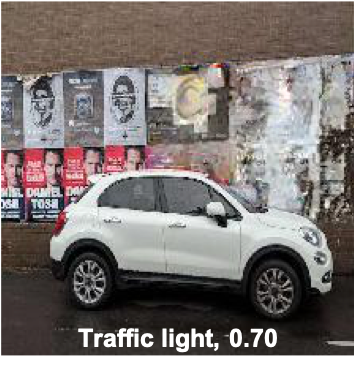}
  \end{subfigure}
  \caption{Camouflaged adversarial images crafted by our \emph{AdvCam} attack and their original versions. }
  \label{fig:use_examples}
\end{figure*}

\subsubsection{Customized examples}
Here, we show how \emph{AdvCam} can craft extremely stealthy camouflages, especially off-target ones. Figure \ref{fig:use_examples} illustrates a few such examples. The first-row shows on-target camouflage examples, and the second-row shows off-target camouflages, which are crafted by attacking a carefully chosen background area. 
For on-target ones, \emph{AdvCam} generates natural and stealthy perturbation on the surface of the target objects. For off-target ones, in the first off-target example (left two images in the second row), we hide the attack into the price tag to fool a VGG-19 classifier to misclassify a revolver into a toilet tissue. In the second example (middle two images in the second row), \emph{AdvCam} successfully camouflages a blenheim spaniel into a bearskin by adding flowers in the background. In the third example (right two images in the second row),  we camouflage the attack into the wall posters in the background which causes a minivan parked in front of the wall to be misrecognized  as a traffic light. These examples not only demonstrate the stealthiness and flexibility of our \emph{AdvCam} attack, but also indicate that threats to deep learning systems are ubiquitous, and in many cases, may hardly be noticeable even to human observers.

\subsection{Physical-world Attacks}\label{sec:physical-comparison}
We further design three physical-world attacking scenarios to test the camouflage power of our \emph{AdvCam} attack. We also perform AdvPatch and PGD attacks for comparison. For PGD attack, we test different $\epsilon$ in (16/255, 32/255, 64/255, 128/255) and show successful adversarial examples with the least $\epsilon$.  We print out the adversarial patterns on a A3 or A4 paper, then take 20 photos at various viewing angles and distances.

\begin{figure}[htb]
  \centering
\begin{subfigure}[b]{0.30\linewidth}
    \includegraphics[width = \linewidth]{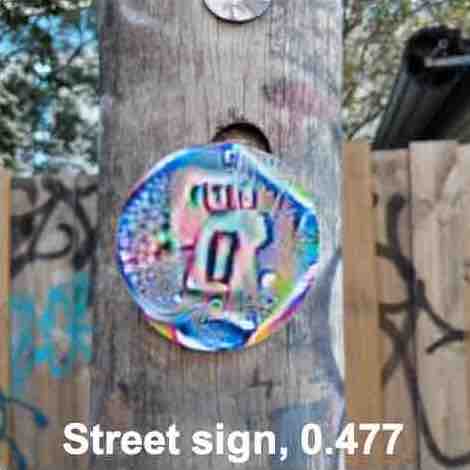}
  \end{subfigure}
  \begin{subfigure}[b]{0.30\linewidth}
    \includegraphics[width = \linewidth]{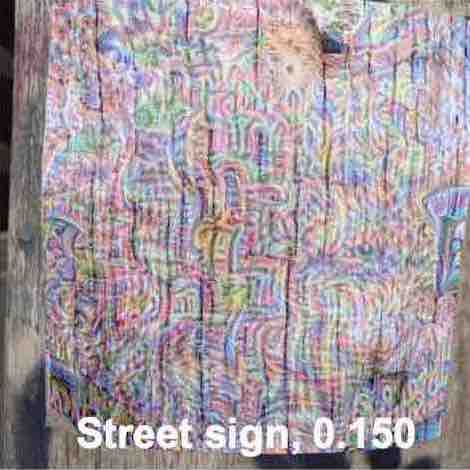}
  \end{subfigure}
  \begin{subfigure}[b]{0.30\linewidth}
    \includegraphics[width = \linewidth]{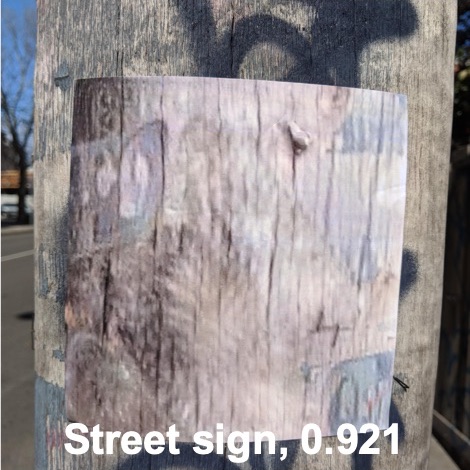}
  \end{subfigure}
  \begin{subfigure}[b]{0.30\linewidth}
    \includegraphics[width = \linewidth]{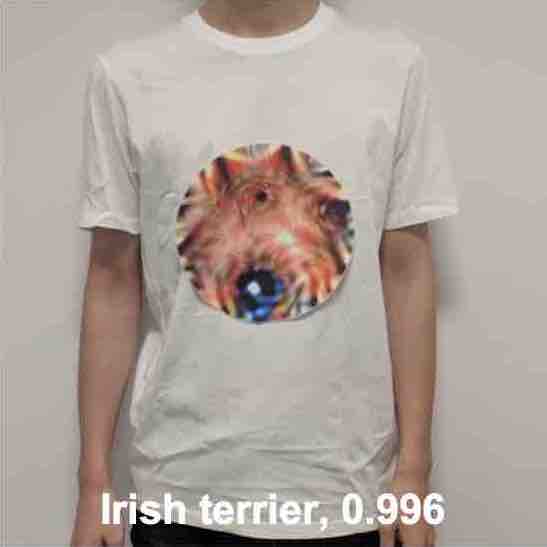}
    \caption{AdvPatch}
  \end{subfigure}
  \begin{subfigure}[b]{0.30\linewidth}
    \includegraphics[width = \linewidth]{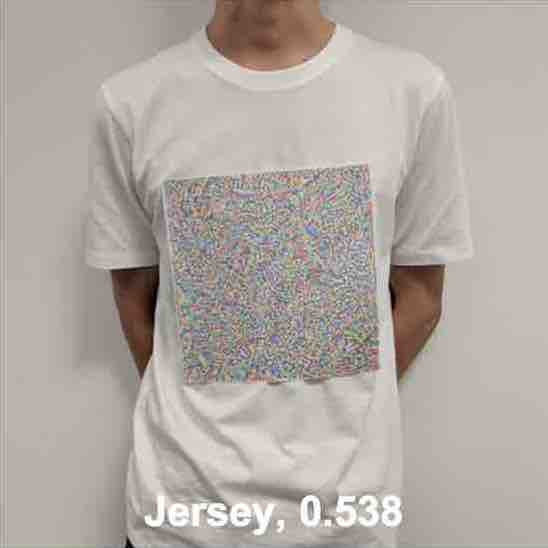}
    \caption{PGD-128}
  \end{subfigure}
  \begin{subfigure}[b]{0.30\linewidth}
    \includegraphics[width = \linewidth]{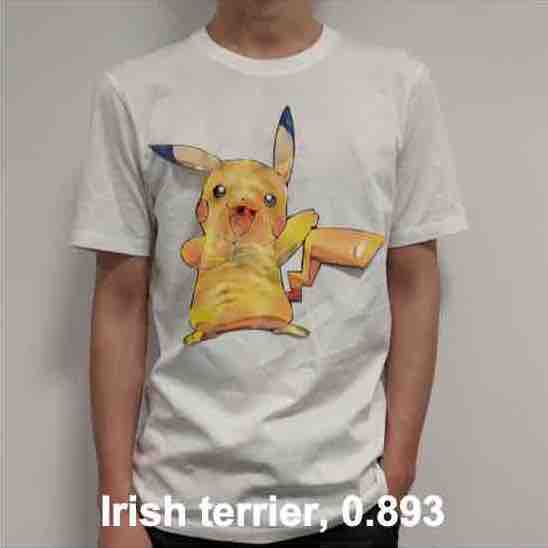}
    \caption{\emph{AdvCam}}
  \end{subfigure}
  \caption{\emph{Top}: Adversarial wood texture recognized as street sign. \emph{Bottom}: Adversarial logo on t-shirt.}
  \label{fig:physical_attack}
\end{figure}

The fist scenario is to camouflage a wild pattern into a street sign, which could cause problems for self-driving cars. The top row in Figure \ref{fig:physical_attack} illustrates some successful patterns crafted by PGD-128 ($\epsilon=128/255$), AdvPatch and \emph{AdvCam}. As can be seen, the attack is perfectly camouflaged by \emph{AdvCam} into the texture of the tree. Although PGD is highly stealthy in digital setting, it requires large perturbation ($\epsilon=128/255$) in physical environment. Thus the adversarial pattern is much less stealthy than \emph{AdvCam}, same as AdvPatch.
The second scenario is to protect the identity of a person wearing a jersey. We simulate such a scenario by attacking the jersey using a camouflaged fashion logo ``pikachu'' (see bottom row in Figure \ref{fig:physical_attack}). All the three attacks perform the ``jersey'' to ``Irish terrier'' attack. Note that PGD failed the attack even with the largest perturbation tested. This shows the high flexibility of \emph{AdvCam} with customized camouflage styles, providing a flexible creation of stealthiness satisfying various attacking scenarios.

\begin{figure}[!t]
  \centering
\begin{subfigure}[b]{0.30\linewidth}
    \includegraphics[width = \linewidth]{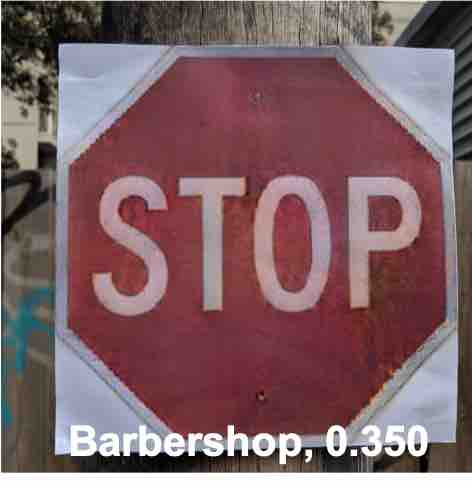}
  \end{subfigure}
  \begin{subfigure}[b]{0.30\linewidth}
    \includegraphics[width = \linewidth]{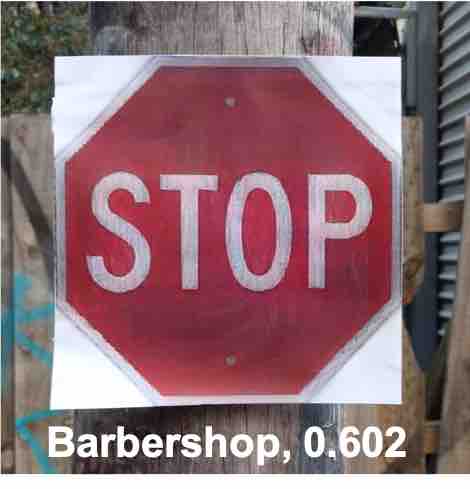}
  \end{subfigure}
  \begin{subfigure}[b]{0.30\linewidth}
    \includegraphics[width = \linewidth]{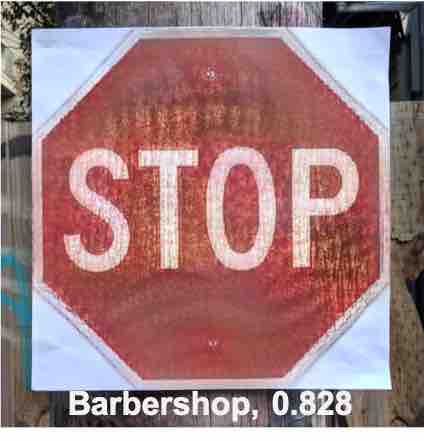}
  \end{subfigure}
  \caption{\label{fig:physical_tra_sign}Adversarial traffic sign with 3 styles of stains.}
  
\end{figure}

We also perform a ``street sign" to ``barbershop" attack using \emph{AdvCam} with three different natural styles (see Figure \ref{fig:physical_tra_sign}). The patterns of \emph{AdvCam} are smooth and natural that can hardly be detected by human observers, but deceive the classifier with high confidence successfully. To summarize, with high stealthiness of adversaries generated by \emph{AdvCam}, it poses ubiquitous threats for current DNNs-based systems. Thus \emph{AdvCam} can be a useful tool to evaluate the robustness of DNNs employed in the physical-world.

\section{Conclusion and Future Work}\label{sec:conclusion}
In this paper, we have investigated the stealthiness of adversarial examples, and propose a novel approach called adversarial camouflage (\emph{AdvCam}), which combines neural style transfer and adversarial attack techniques, to craft and camouflage adversarial examples into stealthy natural-looking styles. 
\emph{AdvCam} is a flexible approach that  can help craft stealthy attacks for robustness evaluation of DNN models. Apart from the view of attack, the proposed \emph{AdvCam} can be a meaningful camouflage technique to protect objects or human being detected by both human observers and DNN based equipments.

The proposed \emph{AdvCam} currently still requires the attacker to manually specify the attack region and target style, where we plan to explore semantic segmentation techniques to automatically achieve this in our future work. 
Also, we will explore to apply \emph{AdvCam} on other computer vision tasks including object detection and segmentation. Moreover, effective defense strategies against camouflaged attacks will be another crucial and promising direction.


\section*{Acknowledgement}
Yun Yang is supported by Australian Research Council Discovery Project under Grant DP180100212. We are also grateful for early stage discussions with Dr Sheng Wen from Swinburne University of Technology.

{\small
\bibliographystyle{ieee_fullname}
\bibliography{AdvCam}
}

\end{document}